\documentclass[letterpaper, 10 pt, conference]{ieeeconf}  

\IEEEoverridecommandlockouts                              
\makeatletter
\let\NAT@parse\undefined
\makeatother

\usepackage{times}
\usepackage{helvet}
\usepackage{courier}
\usepackage{amsmath}
\usepackage{float}
\usepackage{graphicx}
\usepackage{subfigure}
\usepackage{multirow}
\usepackage{algorithm}
\usepackage{algorithmic}
\usepackage[numbers]{natbib}
\usepackage{wrapfig}
\usepackage{color}

\renewcommand{\cite}{\citep}

\belowdisplayskip \abovedisplayskip
\renewcommand{\baselinestretch}{0.98}

\newenvironment{packed_item}{
\begin{itemize}
  \setlength{\itemsep}{1pt}
  \setlength{\parskip}{0pt}
  \setlength{\parsep}{0pt}
}{\end{itemize}}

\newlength{\sectionReduceTop}
\newlength{\sectionReduceBot}
\newlength{\subsectionReduceTop}
\newlength{\subsectionReduceBot}
\newlength{\abstractReduceTop}
\newlength{\abstractReduceBot}
\newlength{\captionReduceTop}
\newlength{\captionReduceBot}
\newlength{\subsubsectionReduceTop}
\newlength{\subsubsectionReduceBot}

\newlength{\horSkip}
\newlength{\verSkip}

\newlength{\figureHeight}
\title{\LARGE \bf
Unstructured Human Activity Detection from RGBD Images
}

\author{Jaeyong Sung, Colin Ponce, Bart Selman and Ashutosh Saxena
\thanks{Jaeyong Sung, Colin Ponce, Bart Selman and Ashutosh Saxena are with the Department of Computer Science, Cornell University, Ithaca, NY.
        {\tt\small js946@cornell.edu, \{cponce,selman,asaxena\}@cs.cornell.edu}}%
}

\begin{document}

\maketitle
\thispagestyle{empty}
\pagestyle{empty}

\begin{abstract}
Being able to detect and recognize human activities is essential for several 
applications, including personal assistive robotics. 
In this paper, we perform detection and recognition of unstructured human activity in
 unstructured environments. We use a RGBD sensor (Microsoft Kinect) as the 
input sensor, and 
compute a set of features based on human pose and motion,
as  well as based on image and point-cloud information.
Our algorithm is based on a hierarchical maximum entropy Markov model (MEMM),
which considers a person's activity as composed of a set of
sub-activities. We infer the two-layered graph structure using a dynamic
programming approach.
We test our algorithm on detecting and recognizing twelve different activities performed
by four people in different environments,
such as a kitchen, a living room, an office, etc., and achieve good performance
even when the person was not seen before in the training set.%
\footnote{
A preliminary version of this work was presented 
at AAAI workshop on Pattern, Activity and Intent Recognition, 2011. 
}
\end{abstract}

\section{Introduction}

Being able to automatically infer the activity that a person is performing is essential
in many applications, such as in personal assistive robotics.
For example, if a robot could watch and keep track of how often a person drinks water,
it could prevent the dehydration of elderly by reminding them.
True daily activities do not happen in structured environments (e.g., with closely controlled 
background), but in uncontrolled and cluttered households and offices.
Due to its unstructured and often visually confusing nature, detection of daily activities becomes a 
much more difficult task. In addition, each person has his or her own habits and mannerisms in 
carrying out tasks, and these variations in speed and style create additional difficulties in trying to 
detect and recognize activities. In this work, we are interested in reliably detecting daily activities 
that a person performs in a home or office,  such as cooking, drinking water, brushing teeth, talking 
on the phone, and so on.


Most previous work on 
activity classification has focused on using 2D video (e.g., \cite{NHWLH,GuptaEfros})
or RFID sensors placed on humans and objects (e.g., \cite{WOCPR}). The use 
of 2D videos leads to relatively low accuracy (e.g., 78.5\% in \cite{LAS}) even when 
there is no clutter.
The use of RFID tags is generally too intrusive because it requires a placement of
RFID tags on the people.


\begin{figure}[tb]
    \centering
    \includegraphics[width=0.48\textwidth]{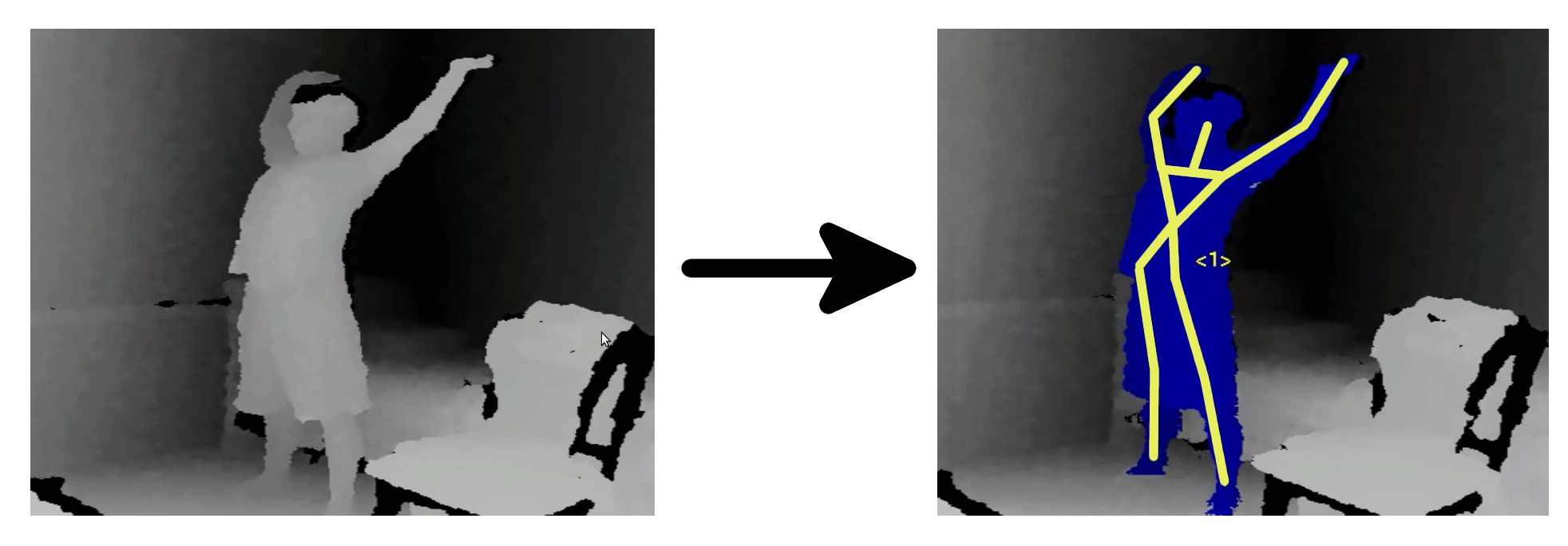}
\vskip -.18in
    \caption{{\small The RGBD data from the Kinect sensor is used to generate an articulated skeleton model. This skeleton is used along with the raw image and depths for estimating the
    human activity.}}
\vskip -.16in
    \label{fig:depthAndSkeleton}

\end{figure}

In this work, we perform activity detection and recognition using an inexpensive RGBD sensor
(Microsoft Kinect).
Human activities, despite their unstructured nature, tend to have a natural hierarchical structure; for instance, drinking water involves a three-step process of bringing a glass to one's mouth, tilting the glass and head to drink, and putting the glass down again. We can capture this hierarchical nature using a hierarchical probabilistic graphical model---specifically, a two-layered maximum entropy Markov model (MEMM). Even with this structured model in place, different people perform tasks at different rates, and any single graphical model will likely fail to capture this variation. To overcome this problem, we present a method of on-the-fly graph structure selection that can automatically adapt to variations in task speeds and style. Finally, 
we need features that can capture meaningful characteristics of the person. We accomplish this by using the PrimeSense skeleton tracking system \cite{PRIMESENSE} in combination with specially placed Histogram of Oriented Gradient \cite{DT} computer vision features. This approach enables us to achieve reliable performance in detection and recognition of common activities performed in typical cluttered human environments.

We evaluated our method on twelve different activities (see Figure~\ref{fig:rgb_images}) performed by four different 
people in five different environments: kitchen, office, bathroom, living room and bedroom. Our results
show a precision/recall of 84.7\%/83.2\% in detecting the correct
activity when the person was seen before in the training set and 67.9\%/55.5\% when the person
was not seen before.
We have also made the dataset and code available open-source at:
 \texttt{http://pr.cs.cornell.edu/humanactivities}





\section{Related Work}
\vspace*{\sectionReduceBot}

There is a large body of previous work on human activity recognition.
One common approach is to use space-time features to model points
of interest in video \cite{Laptev,DRCB}. Several authors have supplemented
these techniques by adding more information to these features
\cite{JSWP,WKC,WOCPR,LAS,CCF,SWT}. However, this approach is only capable of classifying,
rather than detecting, activities.
Other approaches 
include filtering techniques \cite{RAS} and sampling of video patches
\cite{BI}.
Hierarchical techniques for activity recognition have been used as well, but
these typically focus on neurologically-inspired visual cortex-type models
\cite{GP,SWP,ML,RHBL}. Often, these authors adhere faithfully to the models of
the visual cortex, using motion-direction sensitive ``cells'' such as Gabor
filters in the first layer \cite{JSWP,NHWLH}.

Another class of techniques used for activity recognition is that of the hidden
Markov model (HMM). Early work by \citet{BOP} utilized coupled HMMs to
recognize two-handed activities. \citet{WBR} used an HMM together with a
3D occupancy grid to model human actions.
\citet{MOHRV} utilized motion templates together with HMMs to recognize
human activities. 
As well as generative models like HMM, \citet{LWYM} employed a discriminative model
which was aided by interaction analysis between people.
\citet{SKZM} used conditional random fields (CRF) and
maximum-entropy Markov models, arguing that these models
overcome some of the limitations presented by HMMs. 
Notably, HMMs create long-term dependencies between observations and tries to model observations,
which are already fixed at runtime.
On the other hand, MEMM and CRF are able to avoid such dependencies and enables
longer interaction among observations.
However, the use of 2D videos leads to relatively low accuracies.

Other authors have worked on hierarchical dynamic Bayesian networks. 
Early work by \citet{WB} extended HMM to parametric HMM for recognizing pointing gestures. \citet{FST2} introduced hierarchical HMM, which was later extended by
\citet{BPV} to a general structure in which each child 
can have multiple parents. 
\citet{TPBV} then developed a hierarchical semi-Markov CRF that could
be used in partially observable settings. 
\citet{FOX} applied hierarchical CRFs to activity recognition but their model requires many GPS traces and is only capable of off-line classification. \citet{WMG} proposed Dual Hierarchical Dirichlet Processes for surveillance of the large area.
Among several others, the hierarchical HMM is the closest
model of these to ours, but does not capture the idea that a single state may connect 
to different parents only for specified periods of time, as our model does.
As a result, none of these models fit our problem of online detection of human activities
in uncontrolled and cluttered environment.
Since MEMM enables longer interaction among observations unlike HMM \cite{SKZM},
the hierarchical MEMM allows us to take new observations and utilize 
dynamic programming to consider them in an online setting.

Various robotic systems have used activity recognition before. \citet{TAHL}
 used  activity recognition in robotic systems to discern 
 aggressive activities in humans. \citet{LWFS} discuss 
the importance of non-verbal communication between human and robot and 
developed a method to recognize simple activities that are nondeterministic in nature,
 while other works have focused on developing robots that
 utilizes activity recognition to imitate human activities \cite{DM,LMM}.
 However, we are more interested here in assistive robots. Assistive robots are 
robots that assist humans in some task. Several types of assistive robots exist,
 including socially assistive robots that interact with another person in a 
non-contact manner, and physically assistive robots, which can physically help people
\cite{F-SM,TTM,NATJXK,CongcongICRA,JLZS,KAJS}.


\section{Our Approach}
\vspace*{\sectionReduceBot}

We use a supervised learning approach in which we collected ground-truth labeled data for  training our model. 
Our input is RGBD images from a  Kinect sensor,  from which we extract
certain features that are fed as input to our learning algorithm. 
We train a two-layered maximum-entropy Markov model which will capture different properties of 
human activities, including their hierarchical nature and the transitions between sub-activities over time. 



\subsection{Features}
\vspace*{\subsectionReduceBot}

We can recognize a person's activity by looking at his  current pose and movement over time,
as captured by a set of features.
The input sensor for our robot is a RGBD camera (Kinect) that gives us an RGB image as well as depths at each pixel. In order to compute the human pose features, we describe a person by a rigid skeleton that can move at fifteen joints (see Figure~\ref{fig:depthAndSkeleton}).  
We extract this skeleton using a tracking system provided by  \citet{PRIMESENSE}.
The skeleton is described by the length of the links and the joint angles. Specifically, we have the three-dimensional Euclidean coordinates of each joint and the orientation matrix of each joint  with respect to the sensor.  We compute features from this data as follows.


\smallskip
\noindent
\textbf{Body pose features.}
The joint orientation is obtained with respect to the sensor. However, we are interested in true pose, which is invariant of sensor location. Therefore, we transform each joint's rotation matrix so that the rotation is given with respect to the person's torso. 
For 10 joints, we convert each rotation matrix to half-space quaternions in order to more compactly represent the joint's orientation. (A more compact representation would be to use Euler angles, but they suffer from representation problem called gimbal lock \cite{saxena_orientation}.) 
Along with these joint orientations, we would like to know whether person is standing or sitting, and whether or not person is leaning over. Such information is observed from the position of each foot with respect to the torso ($3*2$) by using the head and hip joints to compute the angle of the upper body against vertical.
We have $10*4+3*2+1= 47$ features for the body pose.

\smallskip
\noindent
\textbf{Hand Position.}
Hands play an especially important role in carrying out many activities, so information about what hands are doing can be quite powerful.  In particular, we want to capture information such as ``the left hand is near the stomach'' or ``the right hand is near the right ear.''  To do this, we compute the position of the hands with respect to the torso, and with the respect to the head in the local coordinate frame. Though we capture the motion information as described next, in order to emphasize hand movement, we also observe hand position over last 6 frames and record the highest and lowest vertical hand position.  We have $2*(6+2)=16$ features for this.

\smallskip
\noindent
\textbf{Motion Information.}
Motion information is also important for classifying a person's activities. We select nine frames spread out over the last three seconds, spaced as follows:  $\{ -5, -9, -14, -20, -27, -35, -44, -54, -65\}$, where the numbers refer to the frames chosen. Then, we compute the joint rotations that have occurred between each of these frames and the current frame, represented as half-space quaternions (for the 11 joints with orientation information). This gives. $9*11*4=396$ features. We refer to body pose, hand and motion features as ``skeletal features''.


\smallskip
\noindent
\textbf{Image and point-cloud features.}
Much useful information can be derived directly from the raw image and point cloud as well. We use the Histogram of Oriented Gradients (HOG) feature descriptors  \cite{DT}, which gives 32 features that count how often certain gradient orientations are seen in specified bounding boxes of an image. Although this computation is typically performed on RGB or grayscale images, we can also view the depth map as a grayscale image and compute the HOG features on that. We have two HOG settings that we use. In the ``simple HOG'' setting, we find the bounding box of the person in the image, and compute RGB and depth HOG features for that bounding box, for a total of 64 features. In the ``skeletal HOG'' setting, we use the extracted skeleton model to find the bounding boxes for the person's head, torso, left arm, and right arm, and we compute the RGB and depth HOG features for each of these four bounding boxes, for a total of 256 features. In this paper's primary result, we use the ``skeletal HOG'' setting.


\subsection{Model Formulation}
\vspace*{\subsectionReduceBot}

Human activity is complex and dynamic, and therefore our learning algorithm should model 
different nuances in human activities, such as the following.

\begin{figure} [t!]
    \centering
    \includegraphics[width=1\linewidth]{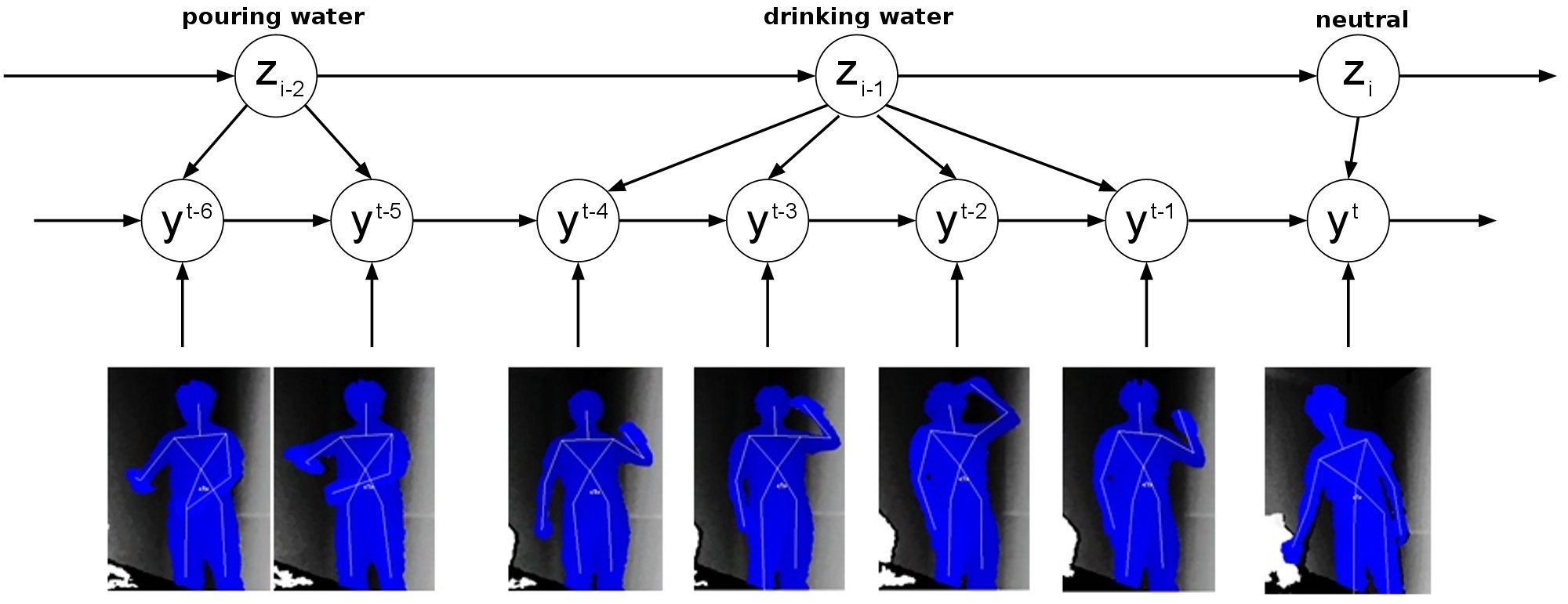}
    \vskip -.15in
    \caption{{\small Our two-layered MEMM model.}}
    \label{fig:memm_gs}
        \vskip -.2in
\end{figure}

First, an activity comprises a series of sub-activities.
For example, the activity ``brushing teeth'' consists of sub-activities such as ``squeezing toothpaste,'' 
``bringing toothbrush up to face,'' ``brushing,'' and so forth. Therefore for each activity
(represented by $z \in Z$), we will model sub-activities (represented by $y \in Y$).
We will train a hierarchical Markov model where the sub-activities $y$ are represented by a layer of hidden variables (see Figure~\ref{fig:memm_gs}).
%


For each activity, different subjects perform the sub-activities for different periods of 
time. It is not clear how to associate the sub-activities
 to the activities. This implies that the graph structure of the model cannot be fixed in advance.
 We therefore determine the connectivity between the $z$ and the $y$ layers in the model
 during inference.

\smallskip
\noindent
\textbf{Model.}
Our model is based on a maximum-entropy Markov model (MEMM) \cite{MEMM}.
However, in order to incorporate the hierarchical nature of activities, we use a
two-layered hierarchical structure, as shown in Figure~\ref{fig:memm_gs}. 


In our model, let $x^{t}$ denote the features extracted from the articulated 
skeleton model at time frame $t$.
Every frame is connected to high-level activities through the mid-level sub-activities. 
Since high-level activities do not change every frame, we do not index them by time. Rather, we simply write $z_i$ to denote the $i^{th}$ high-level activity. Activity $i$ occurs from time $t_{i-1}+1$ to time $t_i$. Then $\{y^{t_{i-1}+1}, ..., y^{t_i}\}$ is the set of sub-activities connected to activity $z_i$.


\subsection{MEMM with Hierarchical Structure}
\vspace*{\subsectionReduceBot}



As shown in Figure \ref{fig:memm_gs}, each node 
$z_i$ in the top layer is connected to several consecutive nodes in the middle layer $\{y^{t_{i-1}+1}, ..., y^{t_i}\}$,
 capturing the intuition that a single activity consists of a number of consecutive sub-activities.

For the sub-activity at each frame $y^{t}$, 
we do not know a priori to which activity $z_i$ it should connect at the top layer. 
Therefore, our algorithm must decide when to connect a middle-layer node $y^t$ to top-layer node $z_i$
and when to connect it to next top-layer node $z_{i+1}$.
We show in the next section how selection of graph structure can be done through dynamic programming.
Given the graph structure, our goal is to infer the $z_i$ that best explains the data. We do this by
modeling the joint distribution $P(z_i, y^{t_{i-1}+1} \cdots y^{t_i} | O_i, z_{i-1})$ where $O_i = x^{t_{i-1}+1}, ..., x^{t_i}$, and for each $z_i$, 
we find the set of $y^t$'s that maximize the joint probability. Finally, we choose the $z_i$ that has the highest 
joint probability distribution.

\textbf{Learning Model.} 
We use a Gaussian mixture model to cluster
the original training data into separate clusters, and consider each cluster as a sub-activity, rather than manually labeling sub-activities for each frame.
We constrain the model to create five clusters for each activity, and then combine all the clusters for a certain location's activities into a single
set of location specific clusters. In addition, we also generate a few clusters from the negative examples, so
that our algorithm becomes robust to not detecting random activities. Specifically, for each classifier and for each
location, we create a single cluster from each of the activities that do not occur in that location. 

Our model consists of the following three terms:
\begin{packed_item}
\item \textbf{$P(y^{t}|x^{t})$:}
This term models the dependence of the sub-activity label $y^t$ on the features $x^t$. 
We model this using the Gaussian mixture model
we have built.   
The parameters of the model are estimated from
the labeled training data using maximum-likelihood.

\item \textbf{$P(y^{t_i - m} | y^{t_i - m - 1}, z_i)$} (where $m \in \{0, ..., (t_i - t_{i-1} - 1)\}$). 
A sequence of sub-activities  describes the activities. 
For example, we can say the sequence ``squeezing toothpaste,'' ``bringing toothbrush up to face,'' 
``actual brushing,'' and ``putting toothbrush down''  describes the activity ``brushing teeth.''
If we only observe ``bringing toothbrush up to face'' and ``putting toothbrush down,'' we would
not refer to it as ``brushing teeth.''
Unless the activity goes through a specific set of sub-activities in nearly the same sequence,
it should probably not be classified as the activity. For all the activities except  \emph{neutral}, 
the table is built from observing the transition of posterior probability for soft cluster of Gaussian mixture model at each frame.

However, it is not so straightforward to build $P(y^{t_i - m} | y^{t_i - m - 1}, z_i)$ when $z_i$ is \emph{neutral}.
When a sub-activity sequence such as ``bringing toothbrush to face'' and ``putting toothbrush down'' occurs,
it does not correspond to any known activity and so is likely to be \emph{neutral}. 
It is not possible to collect data of all sub-activity sequences that
do not occur in our list of activities, so we rely on the sequences observed from non-\emph{neutral} 
activities. If $N$ denotes \emph{neutral} activity, then $P(y^{t_i - m} | y^{t_i - m - 1}, z_i = N) \propto 1-\sum\limits_{z_i \neq N} P(y^{t_i - m} | y^{t_i - m - 1}, z_i).$

\item \textbf{$P(z_i | z_{i-1})$}. 
The activities evolve over time. For example, one activity may be more likely to follow another, 
and there are brief moments of \emph{neutral} activity between
two non-\emph{neutral} activities.
Thus, we can make a better estimate of the activity at the current time
if we also use the estimate of the activity at previous time-step.
Unlike other terms, due to difficulty of obtaining rich data set for maximum likelihood estimation,
 $P(z_i|z_{i-1})$ is set manually to capture these intuitions.

\end{packed_item}

\noindent
\textbf{Inference.}
Consider the two-layer MEMM depicted in Figure \ref{fig:memm_gs}. Let a single $z_i$ activity node 
along with all the $y^t$ sub-activity nodes connected directly to it and the corresponding $x^t$ 
feature inputs be called a \emph{substructure} of the MEMM graph. 
Given an observation sequence $O_i = x^{t_{i-1}+1}, ..., x^{t_i}$ and a previous activity $z_{i-1}$,
 we wish to compute the joint probability $P(z_i, y^{t_{i-1}+1} \dotsb y^{t_i}| O_i, z_{i-1})$:
\begin{align*}
P(z_i&, y^{t_{i-1}+1} \dotsb y^{t_i} | O_i, z_{i-1}) \\
= &P(z_i | O_i, z_{i-1}) P(y^{t_{i-1}+1} \dotsb y^{t_i}| z_i, O_i, z_{i-1}) \\
= &P(z_i | z_{i-1}) 
  \cdot {\displaystyle\prod^{t_i}_{t=t_{i-1}+2}P( y^t|  y^{t-1}, z_i, x^t) } \\
 & \cdot \sum\limits_{y^{t_{i-1}} } P(y^{t_{i-1}+1}| y^{t_{i-1}}, z_i,x^{t_{i-1}+1}) P(y^{t_{i-1}})
\end{align*}
We have all of these terms except $P( y^t|  y^{t-1}, z_i, x^t)$ and $P(y^{t_{i-1}+1}| y^{t_{i-1}}, z_i,x^{t_{i-1}+1})$.
Both terms can be derived as

\begin{align*}
&P( y^t|  y^{t-1}, z_i, x^t) = {P(y^{t-1}, z_i, x^t | y^t ) P(y^t) \over  P(y^{t-1}, z_i, x^t)} \\
\intertext{We make a naive Bayes conditional independence assumption that
        $y^{t-1}$ and $z_i$ are independent from $x^t$ given $y^t$. 
        Using this assumption, 
        we get:}
&P( y^t|  y^{t-1}, z_i, x^t) = {P( y^t |y^{t-1}, z_i) P(y^t | x^t ) \over  P(y^t)}
\end{align*}
%
%
We have fully derived $P(z_i, y^{t_{i-1}+1} \dotsb y^{t_i}| O_i, z_{i-1})$:
\begin{align*}
P(&z_i, y^{t_{i-1}+1} \dotsb y^{t_i}| O_i, z_{i-1}) 
= P(z_i | z_{i-1}) \\
&  \cdot \sum\limits_{y^{t_{i-1}} } {P( y^{t_{i-1}+1}| y^{t_{i-1}}, z_i) P(y^{t_{i-1}+1}|x^{t_{i-1}+1}) \over  P(y^{t_{i-1}+1})}P(y^{t_{i-1}}) \\
& \cdot \displaystyle\prod^{t_i}_{t=t_{i-1}+2} { {P( y^t |y^{t-1}, z_i) P(y^t | x^t ) \over P(y^t)}}  \\
\end{align*}
 Note that this formula can be factorized into two terms where one of them only contains two variables.
\begin{align*}
 P(z_i, y^{t_{i-1}+1} \dotsb y^{t_i}| O_i, z_{i-1}) =  \mathcal{A} \cdot \displaystyle\prod^{t_i}_{t=t_{i-1}+2} \mathcal{B}(y^{t-1},y^t) 
\end{align*}
Because the formula has factored into terms containing only two variables each, this equation can be easily and efficiently optimized. We simply optimize
each factor individually, and we obtain:
\begin{align*}
\max P(z_i&, y^{t_{i-1}+1} \dotsb y^{t_i}| O_i, z_{i-1})= \max\limits_{y^{t_{i-1}+1}} \mathcal{A} \\
& \cdot \max\limits_{y^{t_{i-1}+2}} \mathcal{B}(y^{t_{i-1}+1},y^{t_{i-1}+2}) \dotsb \max\limits_{y^{t_i}} \mathcal{B}(y^{t_i-1},y^{t_i})
\end{align*}

\smallskip
\begin{figure*} [tb]
    \centering
    \includegraphics[width=0.16\textwidth]{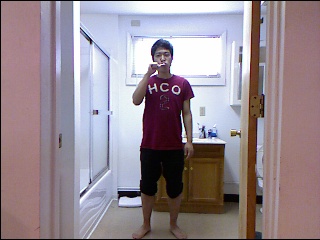}
    \includegraphics[width=0.16\textwidth]{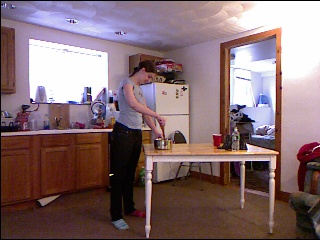}
    \includegraphics[width=0.16\textwidth]{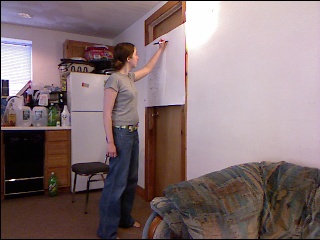}
    \includegraphics[width=0.16\textwidth]{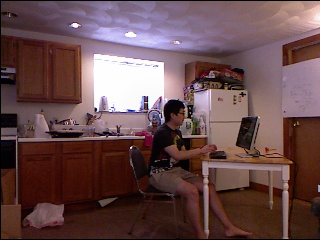}
    \includegraphics[width=0.16\textwidth]{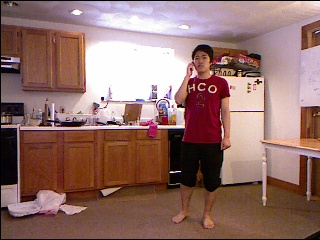}
    \includegraphics[width=0.16\textwidth]{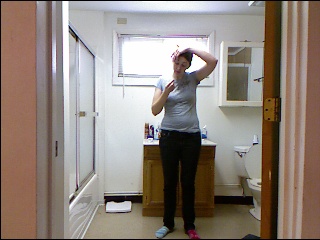}
    \includegraphics[width=0.16\textwidth]{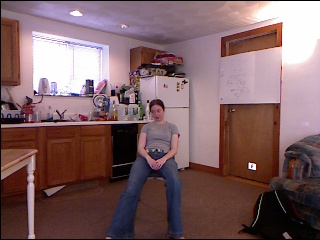}
    \includegraphics[width=0.16\textwidth]{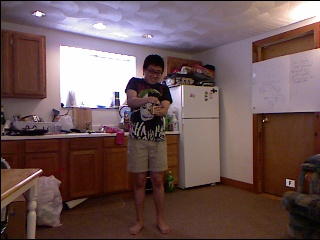}
    \includegraphics[width=0.16\textwidth]{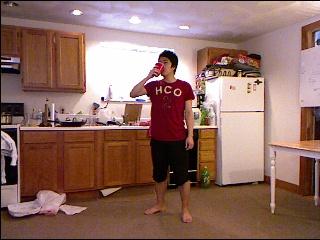}
    \includegraphics[width=0.16\textwidth]{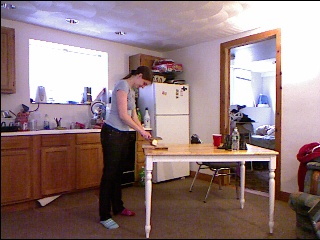}
    \includegraphics[width=0.16\textwidth]{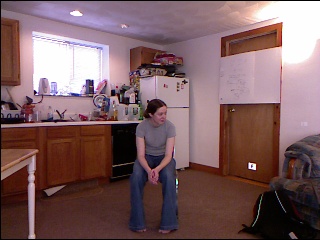}
    \includegraphics[width=0.16\textwidth]{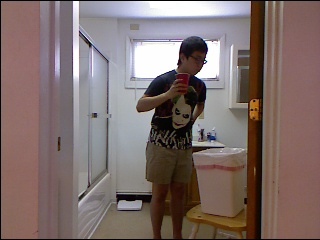}
    \vskip -.13in
    \caption{\small 
        {Samples from our dataset. Row-wise, from left: brushing teeth, cooking (stirring), writing on whiteboard, working on computer, talking on phone, wearing contact lenses, relaxing on a chair,  opening a
    pill container, drinking water, cooking (chopping), talking on a chair, and rinsing mouth with water.}}
    \label{fig:rgb_images}
    \vskip -.1in
\end{figure*}

\subsection{Graph Structure Selection}
\vspace*{\subsectionReduceBot}

\noindent
Now that we can find the set of $y^t$'s that maximize the joint probability $P(z_i, y^{t_{i-1}+1} \dotsb y^{t_i} | O_i, z_{i-1})$,
the probability of an activity $z_i$ being associated with 
the $i^{th}$ substructure and the previous activity, we wish to use that to compute the probability 
of $z_i$ given all observations up to this point. However, to do this, we must solve the following
 problem: for each observation $y^t$, we must decide to which high-level activity $z_i$ it should be connected
 (see Figure \ref{fig:memm_gs}). For example, consider the last $y$ node associated with the ``drinking water''
 activity in Figure \ref{fig:memm_gs}. It's not entirely clear if that node really should connect to
 the ``drinking water'' activity, or if it should connect to the following ``neutral'' activity. Deciding with 
 which activity node to associate each $y$ node is the problem of hierarchical MEMM graph structure selection.

Unfortunately, we cannot simply try all possible graph structures. To see why, suppose we have a
 graph structure at time $t-1$ with a final high-level node $z_i$, and then are given a new node $y^t$.
 This node has two ``choices'': it can either connect to $z_i$, or it can create a new high-level node 
 $z_{i+1}$ and connect to that one. Because every node $y^t$ has this same choice, if we see a total of 
 $n$ mid-level nodes, then there are $2^n$ possible graph structures.

We present an efficient method to find the optimal graph structure using dynamic programming. 
The method works, in brief, as follows. When given a new frame for classification, we try to find 
the point in time at which the current high-level activity started. So we pick a time $t'$, and 
say that every frame after $t'$ belongs to the current high-level activity. We have already computed
 the optimal graph structure for the first $t'$ time frames, so putting these two subgraphs together 
 give us a possible graph structure. We can then use this graph to compute the probability that the 
 current activity is $z$. By trying all possible times $t' < t$, we can find the graph structure that gives us 
 the highest probability, and we select that as our graph structure at time $t$.

\smallskip
\noindent
\textbf{The Method of Graph Structure Selection.} Now we describe the method in detail. Suppose we are at some time $t$; we wish to select the optimal graph structure given everything we have seen so far. We will define the graph structure inductively based on graph structures that were chosen at previous points in time. Let $G_{t'}$ represent the graph structure that was chosen at some time $t' < t$. Note that, as a base case, $G_0$ is always the empty graph.

For every $t' < t$, define a candidate graph structure $\tilde{G}_t^{t'}$ consisting of $G_{t'}$ (the graph structure capturing the first $t'$ timeframes), followed by a single substructure from time $t'+1$ to time $t$ connected to a single high-level node $z_i$. Note that this candidate graph structure sets $t_{i-1} = t'$ and $t_i = t$. Given the set of candidate structures $\{\tilde{G}_t^{t'} | 1 \leq t' < t\}$, the plan is to find the graph structure and high-level activity $z_i \in Z$ to maximize the likelihood given the set of observations so far.

Let $O$ be the set of all observations so far. Then $P(z_i | O ; \tilde{G}_t^{t'})$ is the 
probability that the most recent high-level node $i$ is activity $z_i \in Z$, given all observations 
so far and parameterized by the graph structure $\tilde{G}_t^{t'}$. We initially set $P(z_0 | O; G_0)$ to 
a uniform distribution. Then, through dynamic programming, we have $P(z_{i-1} | O; G_{t'})$ for all $t' < t$ 
and all $z \in Z$ (details below). Suppose that, at time $t$, we choose the graph structure 
$\tilde{G}_t^{t'}$ for a given $t' < t$. Then the probability that the most recent node $i$ is 
activity $z_i$ is given by
\begin{align} \label{eq:gs}
P(z_i | O ; \tilde{G}_t^{t'}) = &\sum_{z_{i-1}} P(z_i, z_{i-1} | O ; \tilde{G}_t^{t'}) \notag \\
= &\sum_{z_{i-1}} P(z_{i-1} | O ; \tilde{G}_t^{t'}) P(z_i | O, z_{i-1} ; \tilde{G}_t^{t'}) \notag \\
= &\sum_{z_{i-1}} P(z_{i-1} | O ; G_{t'}) P(z_i | O_i, z_{i-1})
\end{align}
The two factors inside the summation are terms that we know, the former due to dynamic programming, 
and the latter estimated by finding maximum of $P(z_i, y^{t_{i-1}+1} \dotsb y^{t_i} | O_i, z_{i-1})$, described in the previous section.


Thus, to find the optimal probability of having node $i$ be a specific activity $z_i$, we simply compute
$$
P(z_i | O; G_t) = \max_{t' < t} P(z_i | O ; \tilde{G}_t^{t'})
$$
We store $P(z_i | O ; G_t) \; \forall \, z_i$ for dynamic programming purposes (Equation \ref{eq:gs}). Then, to make a prediction of an activity at time $t$, we compute
\begin{align*}
\mbox{activity}_t = \arg\max_{z_i} P(z_i | O) = \arg\max_{z_i} \max_{t' < t} P(z_i | O ; \tilde{G}_t^{t'})
\end{align*}

\smallskip
\noindent
\textbf{Optimality.} We show that this algorithm is optimal by induction on the time $t$. Suppose we know the optimal graph structure for every time $t' < t$. This is certainly true at time $t = 1$, as the optimal graph structure at time $t=0$ is the empty graph. The optimal graph structure at time $t$ involves a final high-level node $z_i$ that is connected to $1 \leq k \leq t$ mid-level nodes.

Suppose the optimal structure at time $t$ has the high-level node connected to $k = t - t'$ mid-level nodes. Then what graph structure do we use for the first $t'$ nodes? By the induction hypothesis, we  know the optimal graph structure $G_{t'}$ for the first $t'$ nodes. That is, $G_{t'}$ is the graph structure that maximizes the probability $P(z_{i-1} | O)$. Because $z_i$ is conditionally independent of any high-level node before $z_{i-1}$, the graph structure before $z_{i-1}$ does not affect $z_i$. Similarly, the graph structure before $z_{i-1}$ obviously does not depend on the graph structure after $z_{i-1}$. Therefore, the optimal graph structure at time $t$ is $\tilde{G}_t^{t'}$, the concatenation of $G_{t'}$ to a single substructure of $t - t'$ nodes.

We do not know what the correct time $0 \leq t' < t$ is, but because we try all, we are guaranteed to find the optimal $t'$, and therefore the optimal graph structure.

\noindent
\textbf{Complexity.} Let n and m be the number of activities and sub-activities, respectively, and let t be the time.
Space complexity for the dynamic programming algorithm is $O(n\cdot t)$ since we store 1-d array of size t for each activity.
At each timeframe, we must compute the optimal graph structure. By setting a maximum substructure size of $T \ll t$,
dynamic programming requires $n$ activities to be checked for each of $T$ possible sizes.
Each check requires a computation of $P(z_i, y^{t_{i-1}+1} \dotsb y^{t_i} | O_i, z_{i-1})$, which takes $O(m \cdot T)$ time.
Thus, each timeframe requires $O(n\cdot m\cdot T^2)$ computation time.
We do this computation for each of $t$ timeframes, for an overall time complexity of $O(n \cdot m \cdot T^2 \cdot t)$.

\begin{table*}[t] 
{\scriptsize
\renewcommand{\arraystretch}{0.9}
\caption{
    {Results of naive classifier, one-level MEMM model, and our full model in each location.
    The table shows precision and recall scores for all of our models.
    Note that the test dataset contains \emph{random}
    movements (in addition to the activities considered), ranging from a person standing still to walking around while waving his or her hands.
    RGB(D) HOG refers to ``simple HOG''.
    } 
   }
\label{tab:result_table}
\begin{center}
\vskip -0.2in
\setlength{\tabcolsep}{5pt}
\begin{tabular}{@{}c|l@{}|cc|cc|cc|cc|cc|cc|cc|cc@{}}

\hline

\hline
& & \multicolumn{10}{c|}{``New Person''} & \multicolumn{6}{c}{``Have Seen''}\\

& & \multicolumn{2}{c|}{Naive} & \multicolumn{2}{c|}{One-layer} & \multicolumn{6}{c|}{Full Model}& \multicolumn{2}{c|}{Naive} & \multicolumn{2}{c|}{One-layer} & \multicolumn{2}{c}{Full Model}\\
 & & \multicolumn{2}{c|}{Classifier} & \multicolumn{2}{c|}{MEMM}   &\multicolumn{2}{c|}{RGB HOG} & \multicolumn{2}{c|}{RGBD HOG}  & \multicolumn{2}{c|}{Skel.+Skel HOG}&  \multicolumn{2}{c|}{Classifier} & \multicolumn{2}{c|}{MEMM}& \multicolumn{2}{c}{Skel.+Skel HOG} \\
Location & Activity & Prec & Rec  &Prec & Rec &Prec & Rec &Prec & Rec &Prec & Rec & Prec & Rec& Prec & Rec& Prec & Rec\\

\hline

\hline

\multirow{4}{*}{bathroom}
& rinsing mouth    & 77.7 & 49.3 & 71.8 & 63.2 & 42.2 & 73.3 & 49.1 & 97.3 & 51.1 & 51.4 & 73.3 & 49.7 & 70.7 & 53.1 & 61.4 & 70.9\\
& brushing teeth          &  64.5 & 20.5 & 83.3 & 57.7 & 50.7 & 30.8 & 73.4 & 16.6 & 88.5 & 55.3 & 81.5 & 65.1 & 81.5 & 75.6 & 96.7 & 77.1\\
& wearing contact lens    &  82.0 & 89.7 & 81.5 & 89.7 & 44.2 & 40.6 & 52.5 & 59.5 & 78.6 & 88.3 & 87.8 & 71.9 & 87.8 & 71.9 & 79.2 & 94.7\\
\cline{2-18} 
& Average             &  74.7 & 53.1 & 78.9 & 70.2 & 45.7 & 48.2 & 58.3 & 57.8 & 72.7 & 65.0 & 80.9 & 62.2 & 80.0 & 66.9 & 79.1 & 80.9\\
\hline

\multirow{4}{*}{bedroom}
& talking on the phone      &  82.0 & 32.6 & 82.0 & 32.6 & 0.0 & 0.0 & 15.6 & 8.8 & 63.2 & 48.3 & 70.2 & 67.2 & 70.2 & 69.0 &  88.7 & 90.8\\
& drinking water            &  19.2 & 12.1 & 19.1 & 12.1 & 0.0 & 0.0 & 3.0 & 0.1 & 70.0 & 71.7 & 64.1 & 31.6 & 64.1 & 39.6 &  83.3 & 81.7\\
& opening pill container   &  95.6 & 65.9 & 95.6 & 65.9 & 60.6 & 34.8 & 33.8 & 36.5 & 95.0 & 57.4 & 48.7 & 52.3 & 48.7& 54.8 &  93.3 & 77.4\\

\cline{2-18} 
& Average              &  65.6 & 36.9 & 65.6 & 36.9 & 20.2 & 11.6 & 17.4 & 15.2 & 76.1 & 59.2 & 61.0 & 50.4 & 61.0 & 54.5 & 88.4 & 83.3\\
\hline

\multirow{5}{*}{kitchen}
& cooking (chopping)            &  33.3 & 56.9 & 33.2 & 57.4 & 56.1 & 90.0 & 59.9 & 74.2 & 45.6 & 43.3 & 78.9 & 28.9 & 78.9 & 29.0 &  70.3 & 85.7\\
& cooking (stirring)            &  44.2 & 29.3 & 45.6 & 31.4 & 58.0 & 4.0 & 94.5 & 11.1 & 24.8 & 17.7 & 44.6 & 45.8 & 44.6 & 45.8 &  74.3 & 47.3\\
& drinking water               &  72.5 & 21.3 & 71.6 & 23.9 & 0.0 & 0.0 & 91.8 & 23.9 & 95.4 & 75.3 & 52.2 & 51.5 & 52.2 & 52.4 & 88.8 & 86.8\\
& opening pill container        &  76.9 & 6.2 & 75.8 & 6.2 & 83.6 & 33.5 & 54.1 & 35.0 & 91.9 & 55.2 & 17.9 & 62.4 & 17.9 & 62.4 & 91.0 & 77.4\\
\cline{2-18} 
& Average              &  56.8 & 28.4 & 56.6 & 29.7 & 49.4 & 31.9 & 75.1 & 36.1 & 64.4 & 47.9 & 48.4 & 47.2 & 48.4 & 47.4 &  81.1 & 74.3\\
\hline

& talking on the phone     &  69.7 & 0.9 & 83.3 & 25.0 & 0.0 & 0.0 & 31.0 & 11.8 & 51.5 & 48.5 & 34.1 & 67.7 & 34.1 & 67.7 &  88.8 & 90.6\\
living & drinking water    &  57.1 & 53.1 & 52.8 & 55.8 & 0.0 & 0.0 & 1.2 & 0.0 & 54.3 & 69.3 & 80.2 & 48.7 & 71.0 & 53.8 &  80.2 & 82.6\\
room & talking on couch    &  71.5 & 35.4 & 57.4 & 91.3 & 42.7 & 59.4 & 53.2 & 63.2 & 73.2 & 43.7 & 91.4 & 50.7 & 91.4 & 50.7 & 98.8 & 94.7\\
& relaxing on couch      &  97.2 & 76.4 & 95.8 & 78.6 & 0.0 & 0.0 & 100.0 & 21.5 & 31.3 & 21.1 & 95.7 & 96.5 & 95.7 & 96.5 &  86.8 & 82.7\\
\cline{2-18} 
& Average              &  73.9 & 41.5 & 72.3 & 62.7 & 10.7 & 14.9 & 46.4 & 24.1 & 52.6 & 45.7 & 75.4 & 65.9 & 73.1 & 67.2 & 88.7 & 87.7\\
\hline

\multirow{5}{*}{office}
& talking on the phone       &  60.5 & 31.0 & 60.6 & 31.5 & 17.5 & 6.7 & 2.7 & 0.6 & 69.4 & 48.2 & 80.4 & 52.2 & 80.4 & 52.2 &  87.6 & 92.0\\
& writing on whiteboard      &  47.1 & 73.3 & 45.2 & 74.1 & 41.2 & 25.1 & 94.0 & 97.0 & 75.5 & 81.3 & 42.5 & 59.3 & 42.5& 59.3& 85.5 & 91.9\\
& drinking water             &  41.1 & 12.4 & 51.2 & 23.2 & 0.0 & 0.0 & 0.0 & 0.0 & 67.1 & 68.8 & 53.4 &36.7 &53.4 & 36.7& 82.3 & 81.5\\
& working on computer        &  93.5 & 76.8 & 93.5 & 76.8 & 100.0 & 11.9 & 100.0 & 29.0 & 83.4 & 40.7 & 89.2 & 69.3& 89.2& 69.3&  89.5 & 93.8\\
\cline{2-18} 
& Average            &  60.5 & 48.4 & 62.6 & 51.4 & 39.7 & 10.9 & 49.2 & 31.7 & 73.8 & 59.8 &66.4 &54.4 &66.4 &54.4 &  86.2 & 89.8\\
\hline
 
\hline
 \multicolumn{2}{c|}{
 \textbf{Overall Average}} &
\textbf{66.3} &\textbf{41.7} &\textbf{67.2} &\textbf{50.2} &\textbf{33.1} &\textbf{23.5} &\textbf{49.3} &\textbf{33.0} &\textbf{67.9} &\textbf{55.5}  &\textbf{66.4} &\textbf{56.0} &\textbf{65.8} & \textbf{58.1} &\textbf{84.7} &\textbf{83.2}

\\ 
\hline

\hline

\end{tabular}
\end{center}
}
\vskip -.3in
\end{table*}

\section{Experiments}
\vspace*{\sectionReduceBot}

\smallskip
\noindent
\textbf{Data.}
We used the Microsoft Kinect sensor, which outputs an RGB image together with aligned
depths at each pixel at a frame rate of 30Hz.
It produces a 640x480 depth image with a range of 1.2m to
3.5m. The sensor is small enough for it to be mounted on inexpensive mobile
ground robots.

We considered five different environments: office, kitchen, bedroom, bathroom, and living room.
Three to four common activities were identified for each location, giving a total
of twelve unique activities (see Table~\ref{tab:result_table}).
Data was collected from four different people: two males and two females.
None of the subjects were otherwise associated with this project (and hence 
were not knowledgeable of our models and algorithm).
We collected about 45 seconds of data for each activity from each person.
The data was collected in different parts of regular household 
with no occlusion of arms and body from the view of sensor. 
When collecting, the subjects were given basic instructions on how to carry out the activity,
such as ``stand here and chop this onion,'' but were not given any instructions on how the 
algorithm would interpret their movements.
(See Figure~\ref{fig:rgb_images}.)

Our goal is to perform human activity \textit{detection}, i.e., our algorithm must be able 
to distinguish the desired activities from other random activities that people perform.
To that end, we collected \emph{random} activities by asking
the subject to act in a manner unlike any of the previously performed activities.
The \emph{random} activity contains sequence of random movements ranging
from a person standing still to a person walking around and stretching his or her body.
Note that \emph{random} data was only used for testing.

For testing, we experimented with two settings. In the ``new person'' setting, we employed 
leave-one-out cross-validation to test each person's data; i.e. the model was trained on
three of the four people from whom data was collected, and tested on the fourth.
In the other ``have seen'' setting of the experiment, the model was given data about the person 
carrying out the same activity. To achieve this setting, we halved the testing subject's data and included 
one half in the training data set.
So, even though the model had seen the person do the activity at least once,
they had not seen the testing data itself.

Finally, to train the model on both left-handed and right-handed people without needing to film them all,
we simply mirrored the training data across the virtual plane down the middle of the screen.
We have made the data available at:
\texttt{http://pr.cs.cornell.edu/humanactivities/}

\smallskip
\noindent
\textbf{Models.}
We compared two-layered MEMM against two models, naive classifier based on SVM and one-level MEMM.
Both models were trained on full set of features we have described earlier.
\begin{packed_item}
\item \textit{Baseline: Naive Classifier.}
As the baseline model, we used a multi-class support vector machine (SVM) as a way to map features to corresponding activities.
Here SVM is used to map the features to the high-level activities directly.

\item \textit{One-level MEMM.}
This is a one-level MEMM model which builds upon the naive classifier.
$P(y^t|x^t)$ is computed by fitting a sigmoid function to
the output of the SVM. Transition probabilities between activities, $P(y^t|y^{t-1})$, use the 
same table we have built for full model, which in that model is called $P(z_i|z_{i-1})$. Using
$P(y^t|x^t)$ and $P(y^t|y^{t-1})$, we  compute the probability that the person is engaged in
activity $j$ at time $t$.

\item \textit{Hierarchical MEMM.}
We ran our full model with a few different sets of input features in order to show
how much improvement our selection of features brings compared to the set of features
that solely relies on images. 
We tried using ``simple HOG'' features (using a person's full bounding box) 
with just RGB image data, ``simple HOG'' features with both RGB and depth data,
 and skeletal features with the ``skeletal HOG'' features for both RGB and depth data.

\end{packed_item}

\begin{figure*} [tbh]
    \centering
    \subfigure[bathroom] { \includegraphics[width=0.3\textwidth, height=1.05in, trim=30mm 35mm 17mm 10mm]{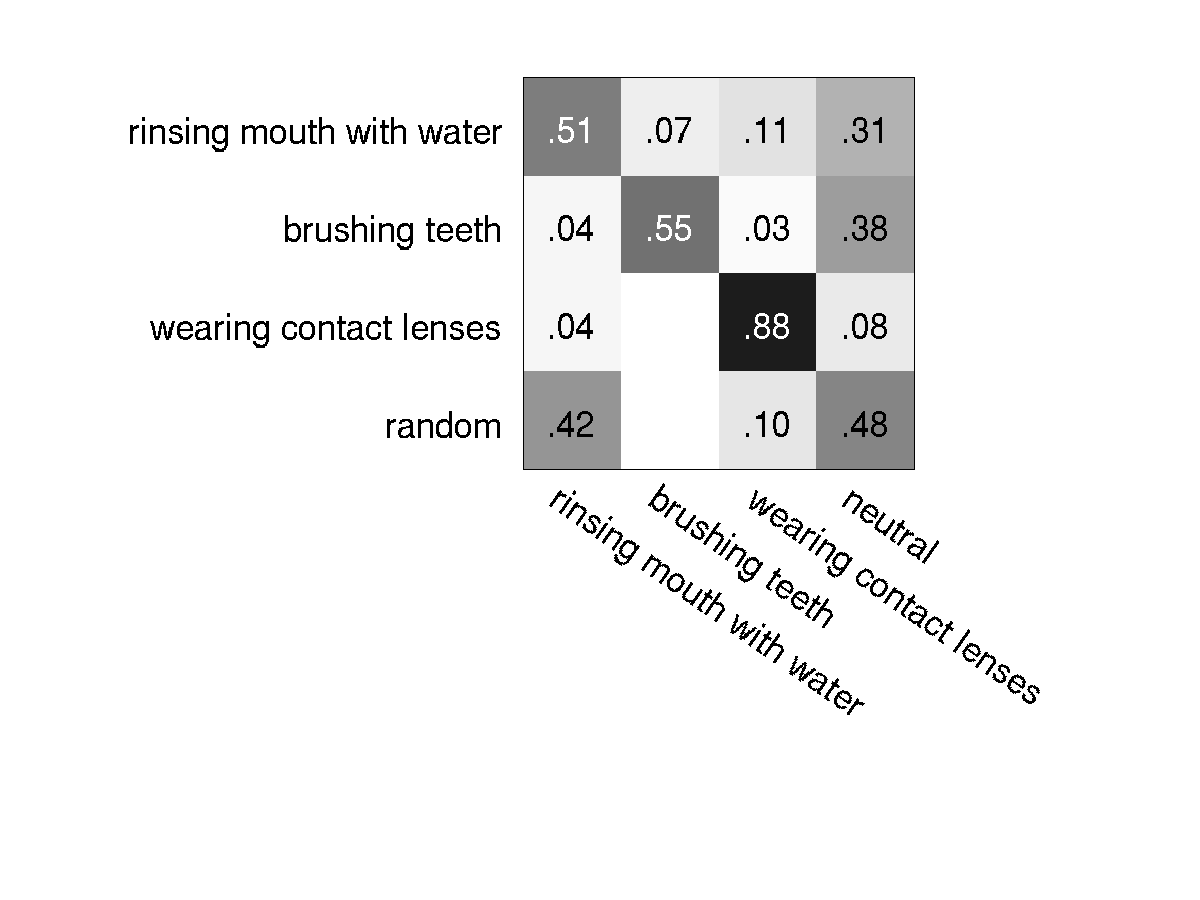} }
    \subfigure[bedroom] { \includegraphics[width=0.3\textwidth, height=1.05in, trim=30mm 27mm 17mm 10mm]{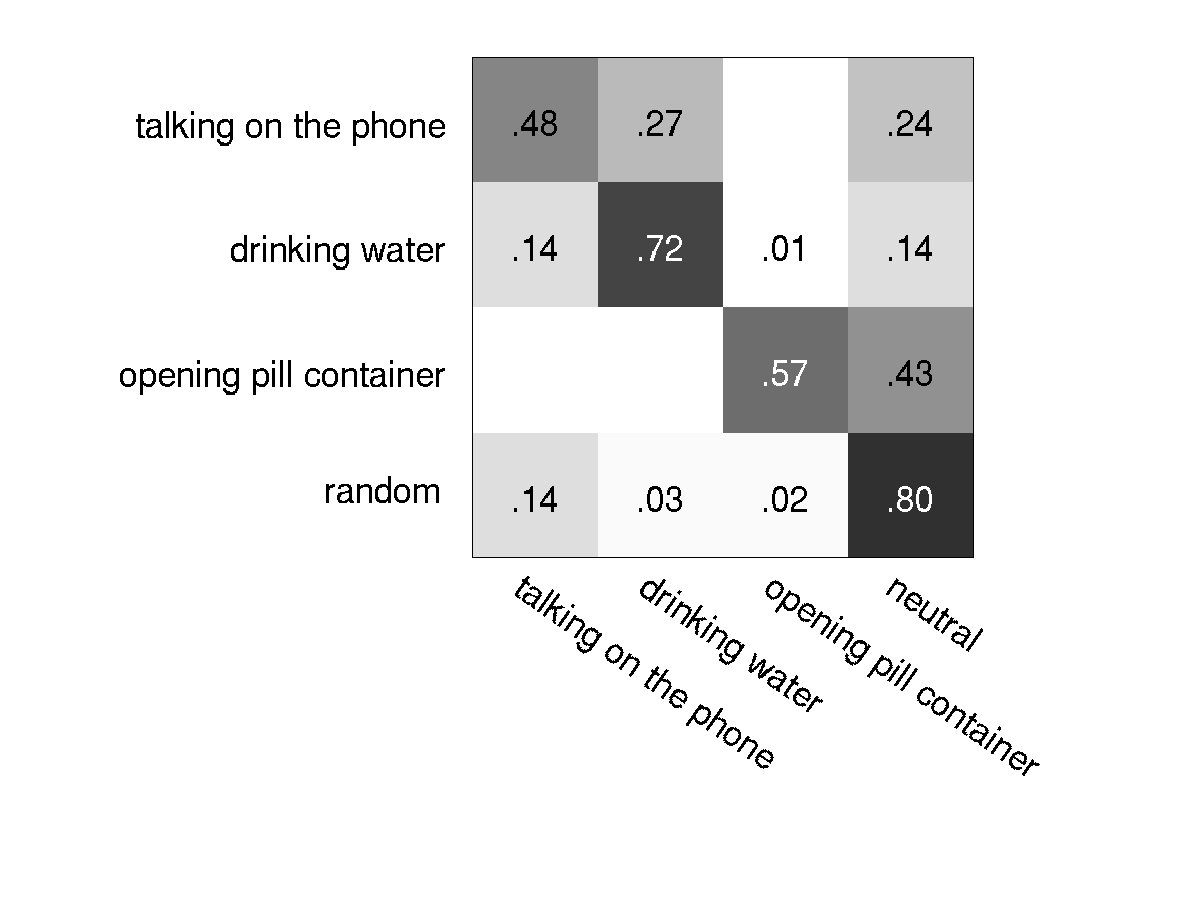}  }
    \subfigure[kitchen] { \includegraphics[width=0.3\textwidth, height=1.05in, trim=30mm 30mm 17mm 10mm]{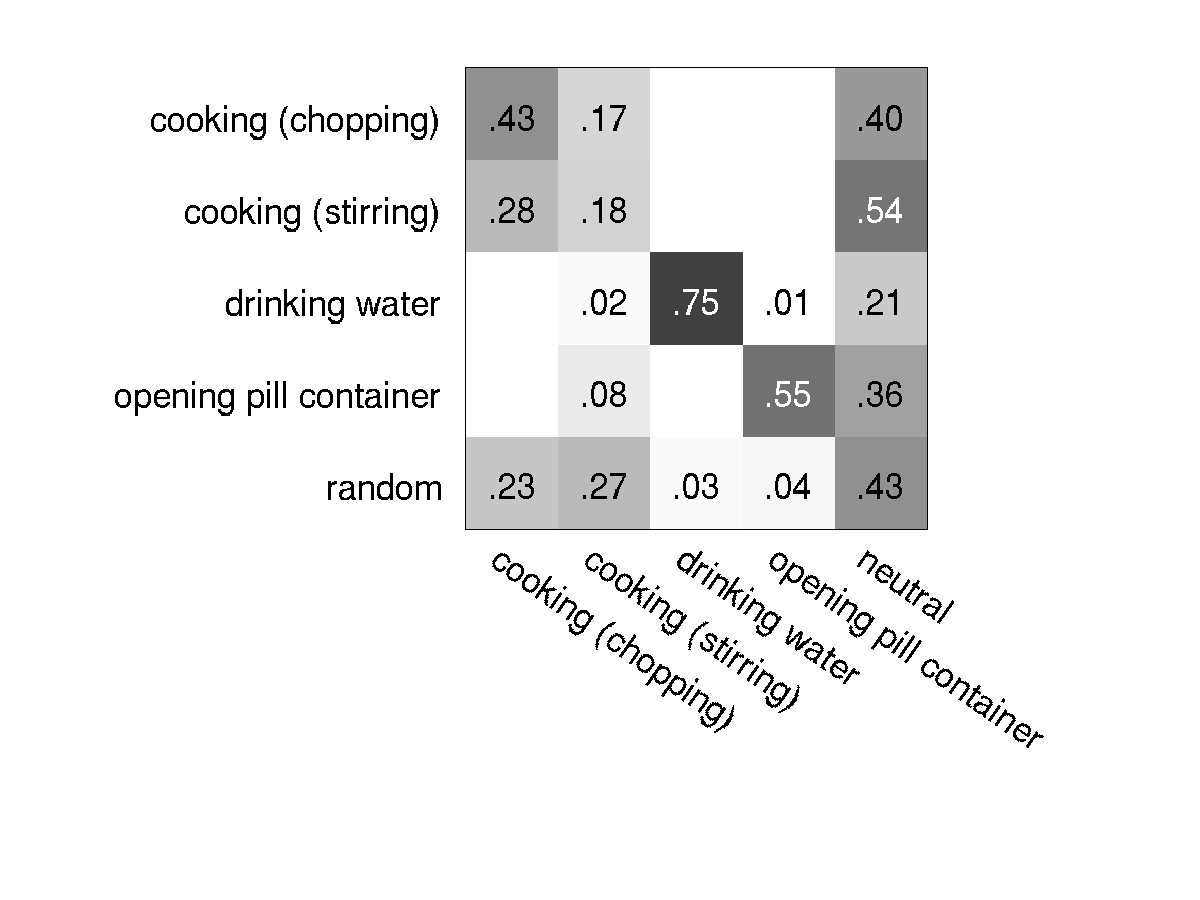}  }
    \subfigure[living room] { \includegraphics[width=0.3\textwidth, height=1.05in, trim=30mm 20mm 12mm 5mm]{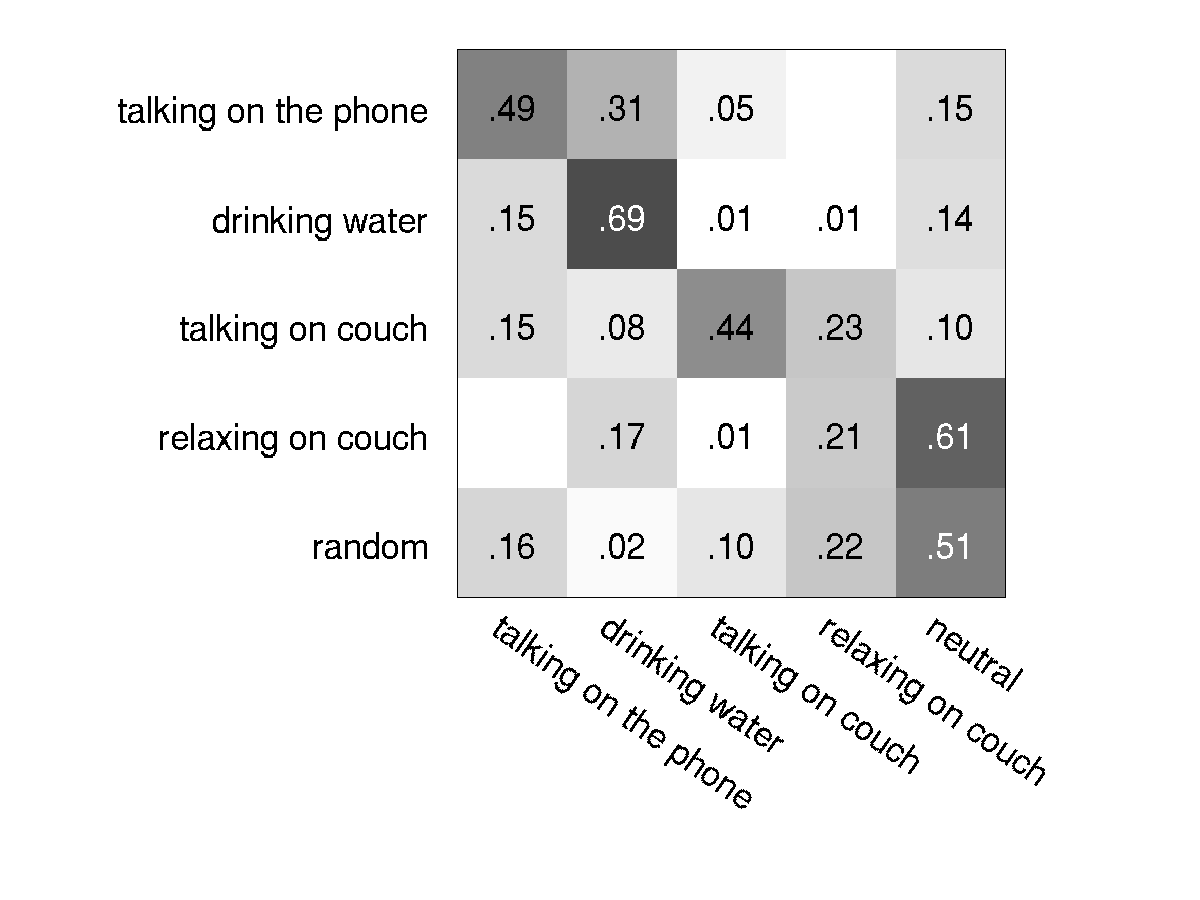} }
    \subfigure[office] { \includegraphics[width=0.3\textwidth, height=1.05in, trim=30mm 20mm 17mm 5mm]{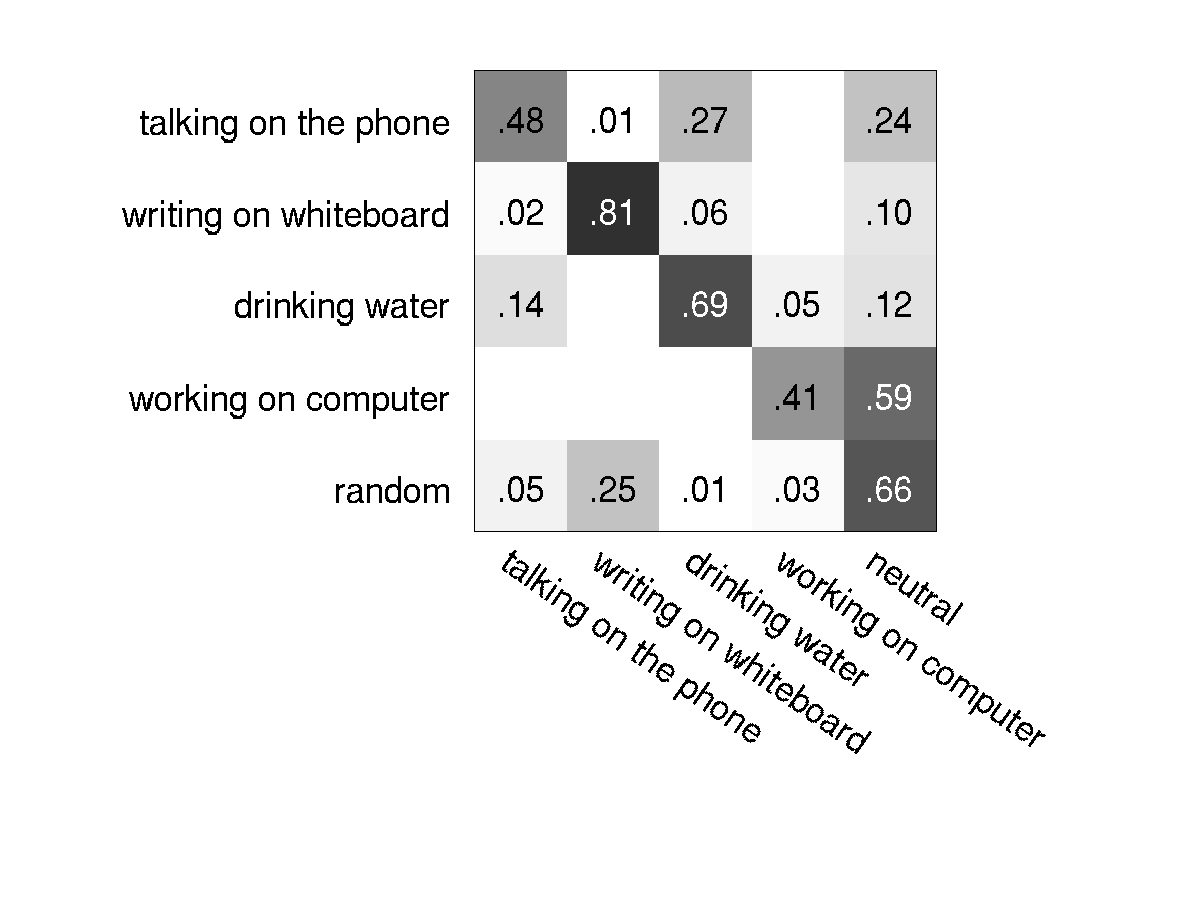} }
    \subfigure[overall] { \includegraphics[width=0.3\textwidth, height=1.05in, trim=30mm 20mm 17mm 5mm]{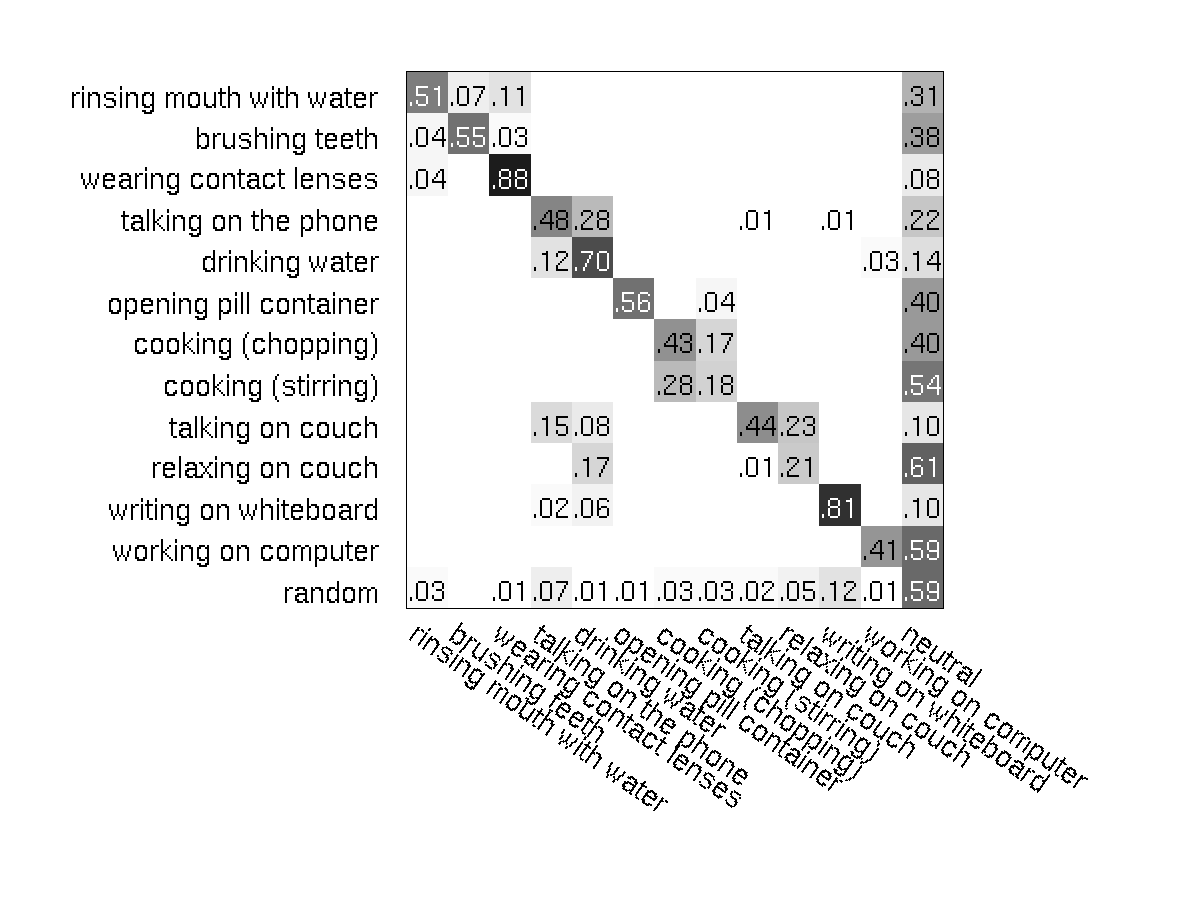}  }
   \vskip -.15in
    \caption{
        \small{Leave-one-out cross-validation confusion matrix for each location with the full model in the ``new person'' setting,
		using skeletal features and skeletal HOG features. The \emph{neutral} activity denotes that the algorithm estimates that the person is either not doing
		anything or that the person is engaged in some other activity that we have not defined. The last matrix (bottom-right)
		shows the results aggregated over all the locations.}}
    \label{fig:confmat_newperson}
\end{figure*}

\begin{figure*} [tbh]
    \centering
    \subfigure[bathroom] { \includegraphics[width=0.3\textwidth, height=1.05in, trim=30mm 35mm 17mm 10mm]{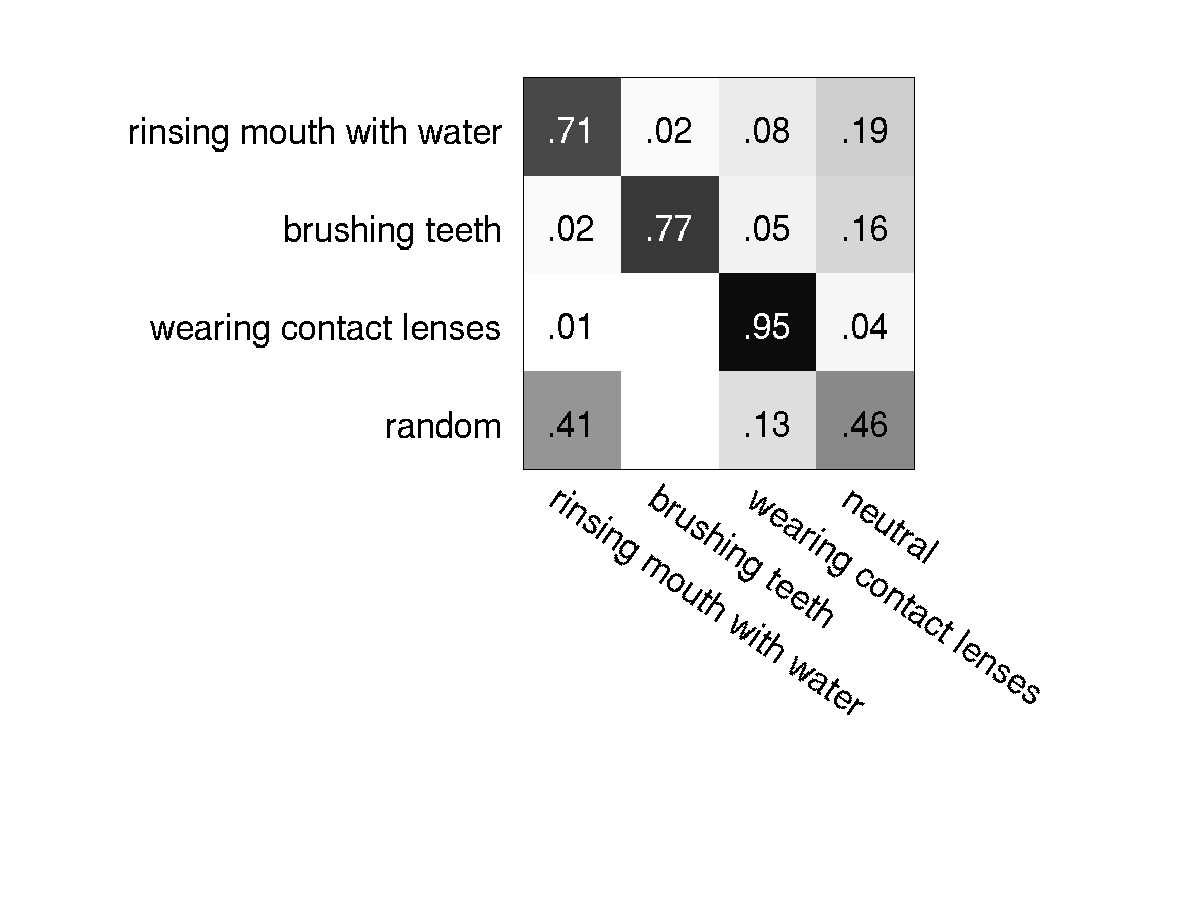} }
    \subfigure[bedroom] { \includegraphics[width=0.3\textwidth, height=1.05in, trim=30mm 27mm 17mm 10mm]{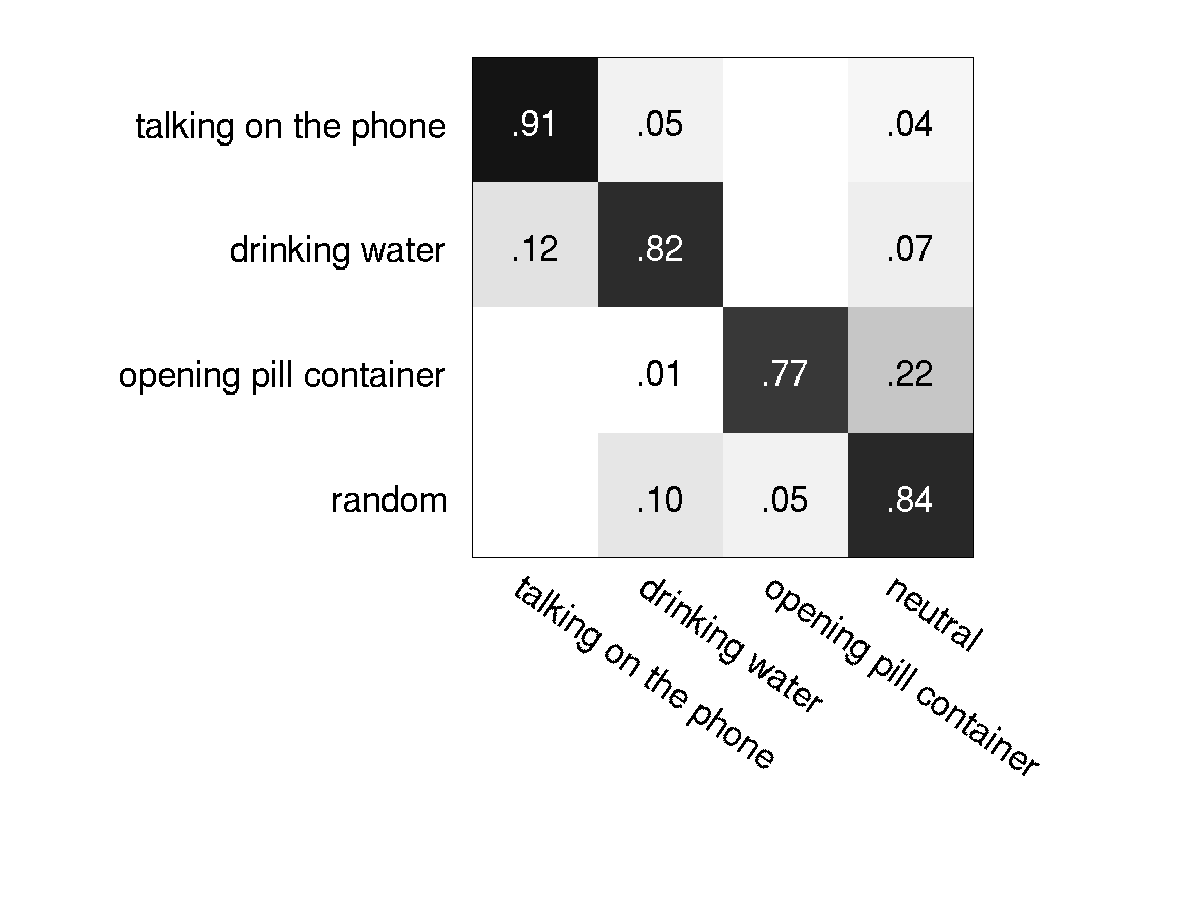}  }
    \subfigure[kitchen] { \includegraphics[width=0.3\textwidth, height=1.05in, trim=30mm 30mm 17mm 10mm]{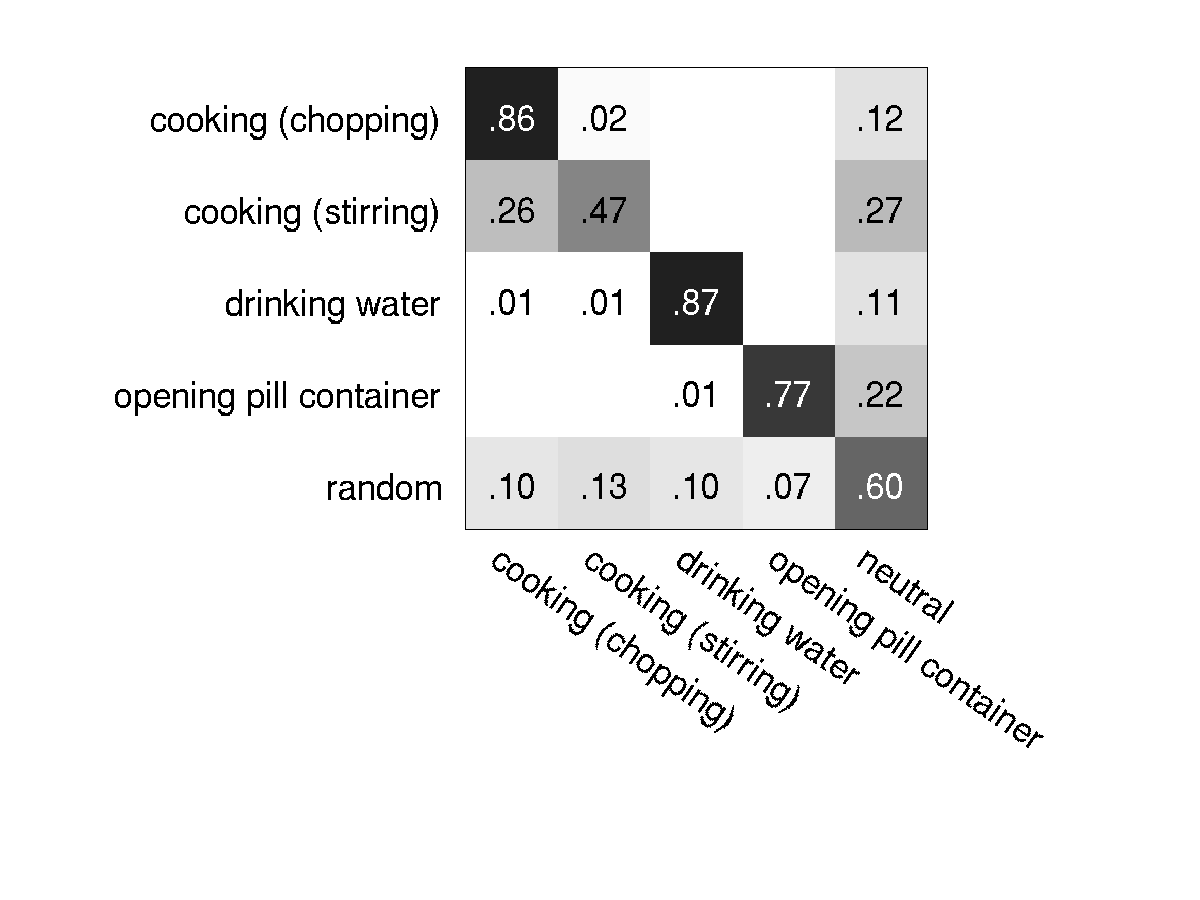}  }
    \subfigure[living room] { \includegraphics[width=0.3\textwidth, height=1.05in, trim=30mm 20mm 12mm 5mm]{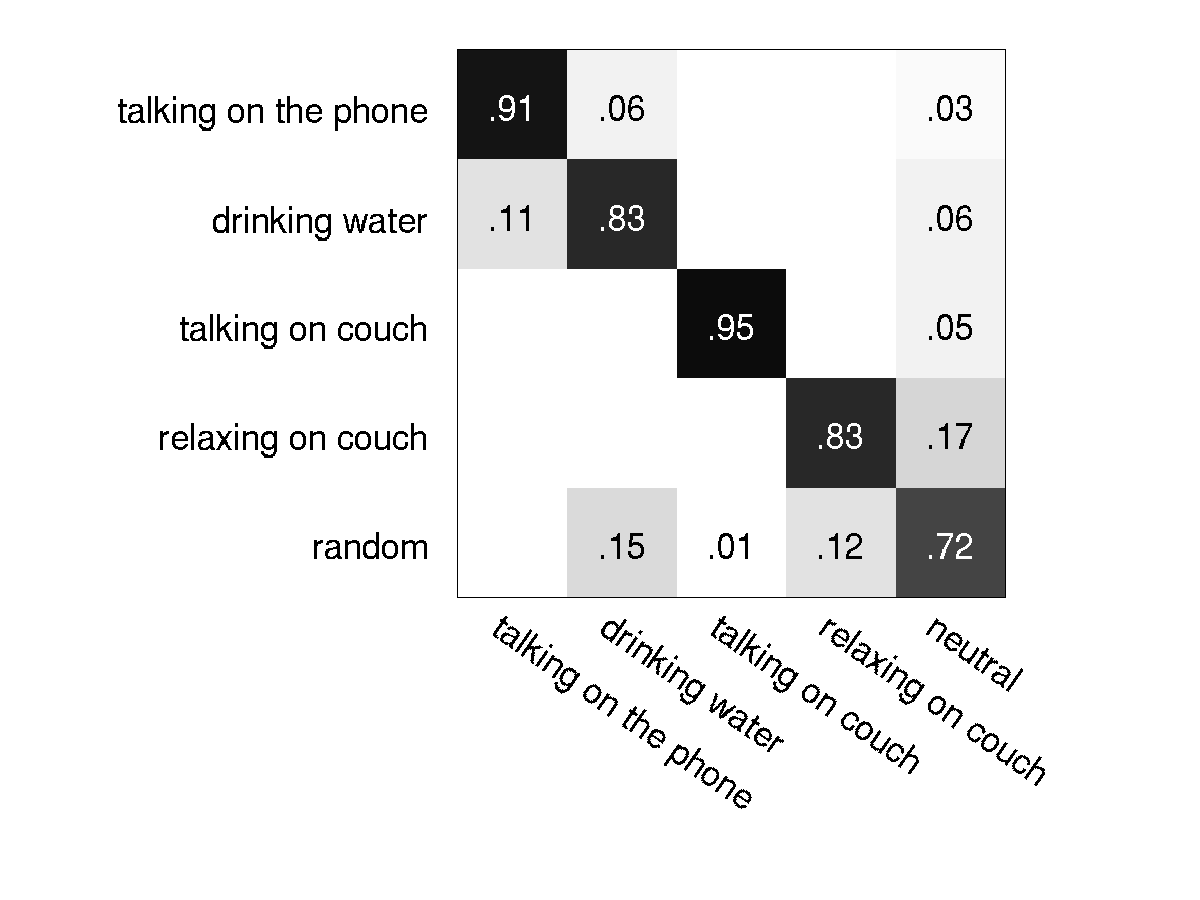} }
    \subfigure[office] { \includegraphics[width=0.3\textwidth, height=1.05in, trim=30mm 20mm 17mm 5mm]{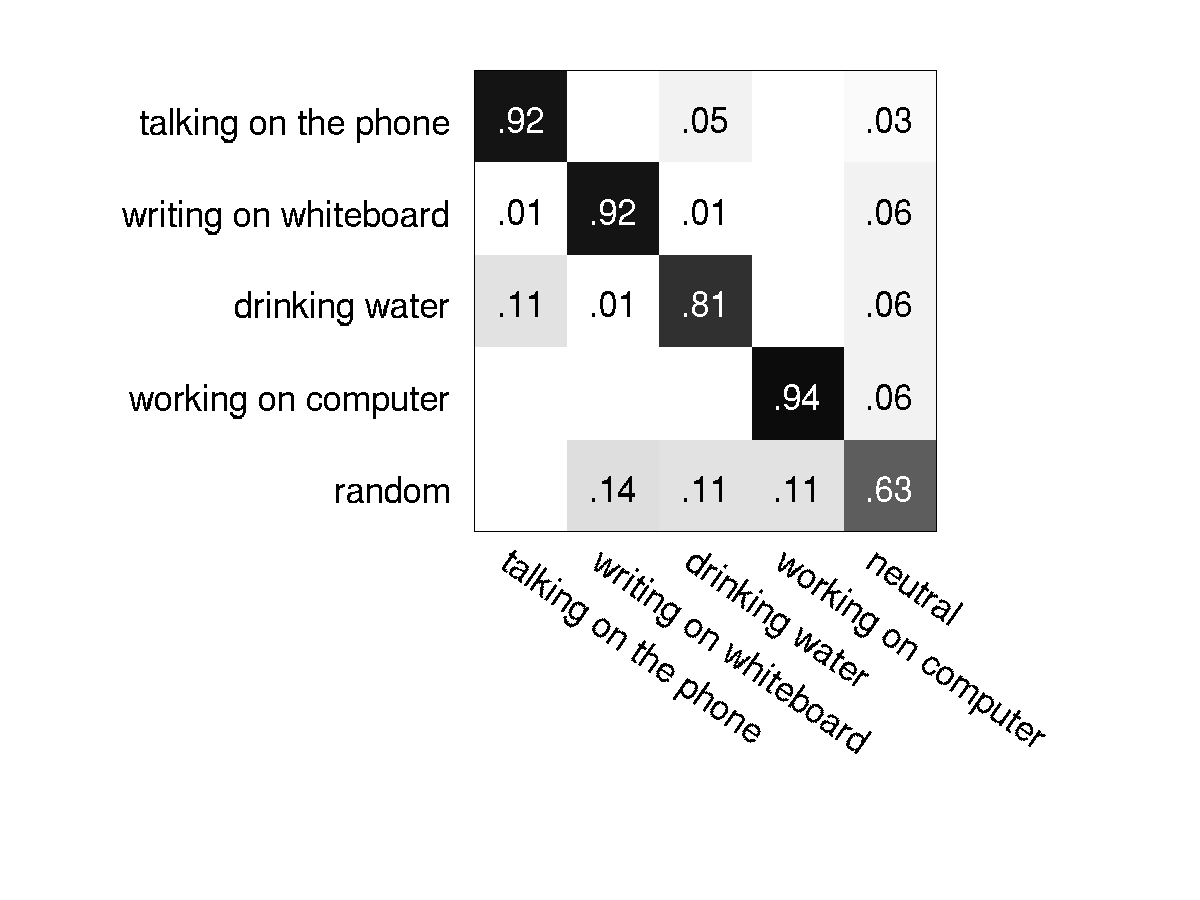} }
    \subfigure[overall] { \includegraphics[width=0.3\textwidth, height=1.05in, trim=30mm 20mm 17mm 5mm]{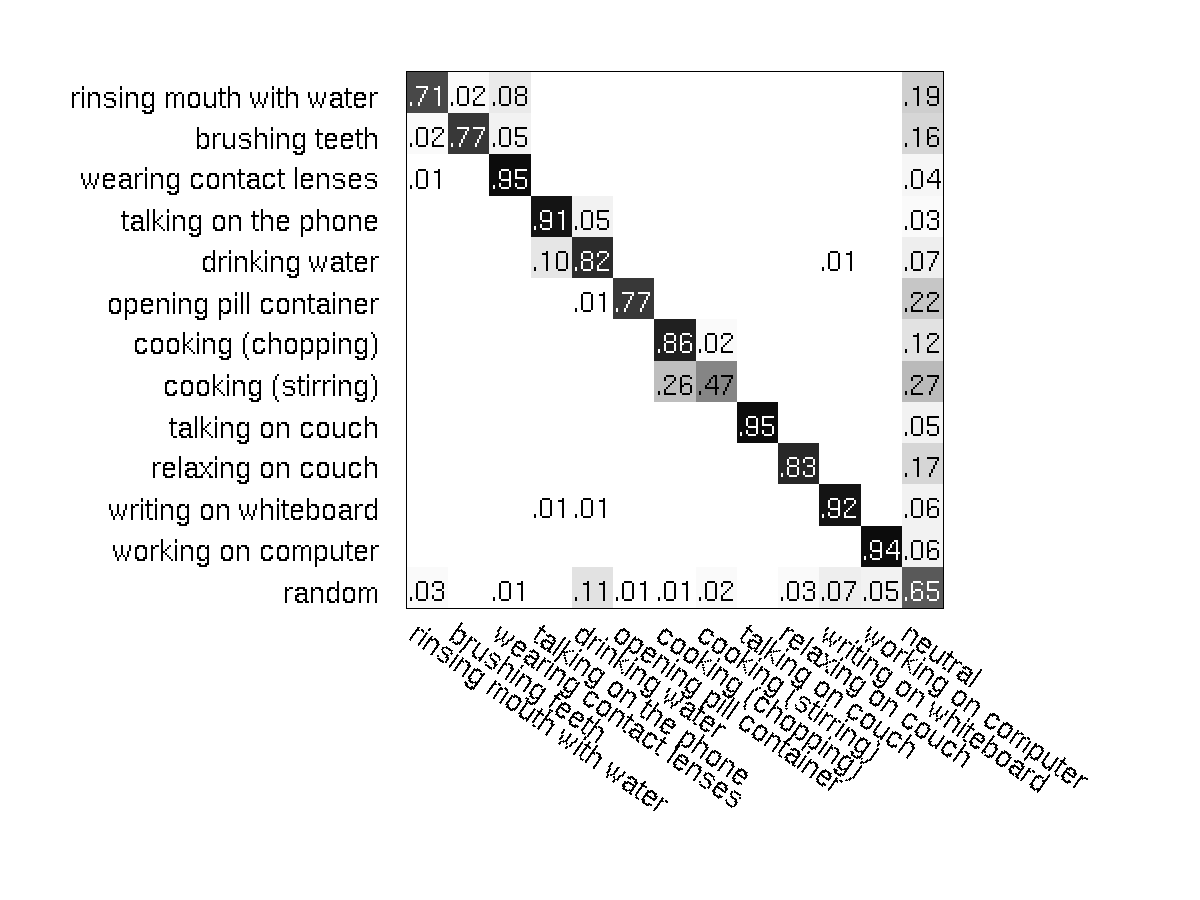}  }
   \vskip -.15in
    \caption{
        \small{Same format as Figure~\ref{fig:confmat_newperson} except it is in the ``have seen'' setting.
		}}
        \vskip -.2in
    \label{fig:confmat_haveseen}
\end{figure*}

\subsection{Results and Discussion}
\vspace*{\subsectionReduceBot}
Table~\ref{tab:result_table} shows the results of the naive classifier, one-level MEMM and our full two-layered model 
for the ``have seen'' and ``new person'' settings. 
The precision and recall measures are used as metrics for evaluation.
Our model was able to detect and classify with a precision/recall measure of 84.7\%/83.2\% and 67.9\%/55.5\%
in ``have seen'' and ``new person'' settings, respectively. It is not surprising that the 
model performs better in the ``have seen'' setting, as it has seen that person's body type and mannerisms before.

We found that both the naive classifier and one-level MEMM were able to classify well
 when a frame contained distinct characteristics of an activity, but performed poorly when characteristics were subtler. The one-layer MEMM was able to perform better than the naive classifier, as it naturally captures important temporal properties of motion. Our full two-layer MEMM, however, is able to capture the hierarchical nature of human activities in a way that neither the naive classifier nor the one-layer MEMM can do. As a result, it performed the best of all three models.

The comparison of feature sets on our full model shows that the features we use
are much more robust compared to features that rely on RGB and/or Depth.

In the ``have seen'' setting,
the HOG on RGB images are capable of capturing powerful information about a person.
However, when seeing a new person, changes in clothing and background can cause confusion especially in 
uncontrolled and cluttered backgrounds,
as shown by relatively low precision/recall value of 33.1\%/23.5\%.
The skeletal features along with HOG on depth, while sometimes less informative than the HOG on images,
are both more robust to changes in people. Thus, by combining skeletal features, 
skeletal HOG image features, and skeletal HOG depth features,
we simultaneously achieved good accuracy in the ``new person'' setting
and very good accuracy in the ``have seen'' setting.


Figure~\ref{fig:confmat_newperson} and Figure~\ref{fig:confmat_haveseen} show the confusion matrices 
between the activities in ``new person'' and ``have seen'' setting when using skeletal features and ``skeletal HOG'' image and depth features.
When it did not classify correctly, it usually chose the \emph{neutral} activity,
which is typically not as bad as choosing a wrong ``active'' activity.
When we look at the confusion matrices, we see that
many of the mistakes are actually reasonable in that the algorithm confuses them
with very similar activities. For example, cooking-chopping and cooking-stirring are often
confused, rinsing mouth with water is confused with brushing teeth, and talking
on the couch is confused with relaxing on the couch.

Another strength of  our model is that it correctly
classifies \emph{random} data as \emph{neutral} most of the time, as shown in the bottom row of the confusion matrices.
This means that
it is able to distinguish whether the provided set of activities actually occurs or not---thus 
our algorithm is not likely to misfire when a person is doing some new activity that the
algorithm has not seen before. 
Also, since we trained on both the regular and mirrored data, the model performs
well with both left- and right-handed people.

However, there are some limitations to our method. 
First, our data only included cases in which the person was not occluded by an object; 
our method does not model occlusions and may not be robust to such situations. 
Second, some activities require more contextual information other than simply human pose.
For example, knowledge of objects being used could help significantly in making 
human activity recognition algorithms more powerful in the future.

\section{Conclusion}
\vspace*{\sectionReduceBot}

In this paper, we considered the problem of detecting and recognizing activities that
humans perform in unstructured environments such as homes and offices.
We used an inexpensive RGBD sensor (Microsoft Kinect) as the 
input sensor, the low cost of which enables our approach to be useful for applications such as 
smart homes and personal assistant robots.
We presented a two-layered maximum entropy Markov model (MEMM). This MEMM modeled 
different properties of the human activities, including their hierarchical nature,
the transitions between sub-activities over time, and the relation between sub-activities
and different types of features. During inference, our algorithm exploited the hierarchical nature of human activities
to determine the best MEMM graph structure.
We tested our algorithm extensively on twelve different activities performed by four different people
in five different environments, where the test activities were often interleaved with random
activities not belonging to these twelve categories. It achieved good detection performance in both settings, where the person was and was not seen before in the training set, respectively.


{
\footnotesize
\bibliographystyle{abbrvnat}
\bibliography{Sung}
}
\end{document}